\documentclass[a4paper,fleqn]{cas-dc}

\usepackage[numbers]{natbib}
\usepackage{algorithm}
\usepackage[noend]{algpseudocode}
\def\tsc#1{\csdef{#1}{\textsc{\lowercase{#1}}\xspace}}
\tsc{WGM}
\tsc{QE}
\tsc{EP}
\tsc{PMS}
\tsc{BEC}
\tsc{DE}

\begin{document}
\let\WriteBookmarks\relax
\def\floatpagepagefraction{1}
\def\textpagefraction{.001}
\shorttitle{Mutual Information Regularized Identity-aware FER in Compressed Video}
\shortauthors{X Liu et~al.}

\title [mode = title]{Mutual Information Regularized Identity-aware Facial Expression Recognition in Compressed Video}

\tnotetext[1]{The funding support from Jiangsu Youth Programme [BK20200238], PolyU Central Research Grant G-YBJW, and Hong Kong Government General Research Fund GRF (Ref. No.152202/14E) are greatly appreciated.}

\author[1]{Xiaofeng Liu}[type=editor,
                        auid=000,bioid=1,
                        orcid=0000-0002-4514-2016]
 
\ead{liuxiaofengcmu@gmail.com}
\ead[url]{https://liu-xiaofeng.github.io/}

\address[1]{Harvard University, Cambridge, MA, USA}
 
\author[1,2]{Linghao Jin} 
 
\address[2]{John Hopkins University, Baltimore, MD, USA}
\ead{ljin23@jhu.edu}

\author[1,2]{Xu Han} 
\ead{xhan32@jhu.edu}

\author[3]{Jane You} 
 \ead{csyjia@comp.polyu.edu.hk}
\ead[url]{https://www.comp.polyu.edu.hk/en-us/staffs/detail/1261}

\address[3]{Dept. of Computing, The Hong Kong Polytechnic University, Hong Kong}

\begin{abstract}
How to extract effective expression representations that invariant to the identity-specific attributes is a long-lasting problem for facial expression recognition (FER). Most of the previous methods process the RGB images of a sequence, while we argue that the off-the-shelf and valuable expression-related muscle movement is already embedded in the compression format. In this paper, we target to explore the inter-subject variations eliminated facial expression representation in the compressed video domain. In the up to two orders of magnitude compressed domain, we can explicitly infer the expression from the residual frames and possibly extract identity factors from the I frame with a pre-trained face recognition network. By enforcing the marginal independence of them, the expression feature is expected to be purer for the expression and be robust to identity shifts. Specifically, we propose a novel collaborative min-min game for mutual information (MI) minimization in latent space. We do not need the identity label or multiple expression samples from the same person for identity elimination. Moreover, when the apex frame is annotated in the dataset, the complementary constraint can be further added to regularize the feature-level game. In testing, only the compressed residual frames are required to achieve expression prediction. Our solution can achieve comparable or better performance than the recent decoded image-based methods on the typical FER benchmarks with about 3 times faster inference.

\end{abstract}


\maketitle

\section{Introduction}

The video modality is increasingly important in many computer vision applications \citep{baddar2019fly,liu2019permutation}. Considering the natural dynamic property of the human face expression \citep{liu2018adaptive}, many works propose to explore spatio-temporal features of facial expression recognition (FER) from the videos. Recently, deep neural networks (DNN) have achieved significant progress for image-based facial expression recognition \citep{liu2019hard}, while the processing of expression video is still challenging.

Although the multi-frame series can inherit richer information and the temporal-correlation between consecutive frames can typically be useful for FER, the video also introduced a lot of redundancy. The signal-to-noise ratio (SNR) in FER videos is exceptionally low due to the slight muscle activity \citep{li2020deep,liu2019hard}.

\begin{figure}[t]
\centering
\includegraphics[width=8cm]{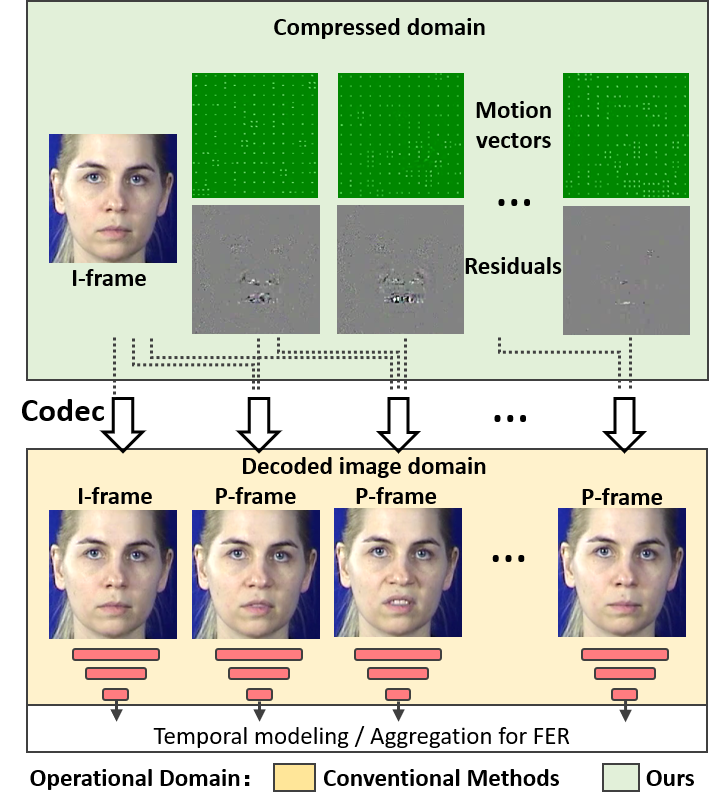}\\
\caption{Illustration of the typical video compression and the scheme of conventional FER methods, which first decode the video and then feed it into a FER network.}\label{fig:1}
\end{figure}

The typical spatio-temporal FER DNNs explore the spatial and temporal clues in a series of frames to extract the facial expression-related information \citep{li2020deep}. The recursive neural network (RNN), 3D convolutional, and non-local networks are the most frequently used network backbones \citep{kim2017multi,meng2019frame}. Unfortunately, using DNN to process many consecutive frames can be computationally costly. It can be hard to scalable for the long videos \citep{liu2017adaptive,liu2019hard}. Furthermore, for some networks, such as RNNs, modeling long-term dependence can be difficult \cite{liu2019permutation}. Many FER techniques, in particular, are able to achieve good performance using image-based FER with decision-level fusion, which completely lost the temporal dependence. This suggests the temporal cues can be hard to explore in the low SNR FER videos with the conventional solutions \citep{kahou2016emonets,xu2016video}.



We argue that the compressed domain can be suitable for the FER task for four reasons. \textbf{1)} the consecutive frames in video modality have many uninformative and repeating patterns, which may drown the ``interesting" and ``true" signal \citep{yeo2006compressed}. With the standard video compression algorithms, the compression ratio can usually be hundred times \citep{richardson2003h264}. Manipulating on the compressed domain can significantly reduce the cost of computation and memory. \textbf{2)} the typical compression methods (e.g., MPEG-4, H.264, and HEVC) break the video to the I frame (intracoded frames) with the first image, and follows several P frames (predictive frames) which encoded as the ``change" or ``movement" \citep{le1991mpeg}, as shown in Figure. \ref{fig:1}. The fundamental of expression is the action of the face muscle.. Many FER systems, in reality, are based on the action unit framework \citep{lucey2010extended}.

As a result, compressed P frames will inherit off-the-shelf but useful expression-related factors, and their pattern is substantially simpler than raw images. \textbf{3)} our compressed domain exploration can also be effective since it focuses on the ``true" signals rather than processing the repeatedly near-duplicates \citep{wu2018compressed}. \textbf{4)} Since the to-be-processed data is transmitted in compressed format, the decoding procedure is not needed in the real-world mission.

Furthermore, the FER task has long suffered from the high inter-subject variation caused by identity discrepancies in facial attributes \citep{meng2017identity,liu2017adaptive}. The learned features may capture more identity-related information than expression-related information, and are not purely related to the FER task. Noticing that the P frames may contain the relative location of face key points, which can be related to the identity \citep{calder2005understanding}. Metric learning is a standard approach for extracting identity from the expression representation \citep{meng2017identity,liu2017adaptive,liu2019hard}. Inspired by the adversarial disentanglement, \citep{ali2019all,cai2019identity} propose to render the identity removed face, which is inspired by adversarial disentanglement (GAN). These researchs, on the other hand, concentrate on image-based FER. \citep{liu2018adaptive} extend the metric learning \citep{liu2017adaptive} for video data by explicitly substitute the image with the video features, without taking into account the video's characteristics. Furthermore, these methods necessitate the use of the identity label and multiple expressions of the same person, which significantly restricts their applicability to the in-the-wild FER task \citep{dhall2014emotion}. 

In this paper, we target to exploit the identity information from the I frame using a pre-trained face recognition network, e.g., FaceNet \citep{schroff2015facenet}. Their identity embeddings are remarkably reliable, since they achieve high accuracy over millions of identities \citep{kemelmacher2016megaface}, and robust to a broad range of nuisance factors such as expression, pose, illumination and occlusion variations.

Using the identity feature as the anchor, we can explicitly enforce the marginal independence of our identity and expression feature. Instead of the complicated adversarial training \citep{liu2019feature} for disentanglement, we adopt the mutual information (MI) as the statistical measure of the independence of these two representations \citep{liu2021mutual}. The MI of two random variables can usually be intractable to directly and precisely measure in a high-dimensional space \citep{linsker1988self}. Recently, some of the works illustrate that mutual information can be differentiable approximated \citep{belghazi2018mine}. We propose to minimize the differentiable MI measure as the objective. Practically, it can be a latent space min-min game of an encoder-discriminator framework, which follows a collaborative fashion rather than adversarial competition. We note that GAN is notorious for its unstable model collapse \citep{goodfellow2016nips}, while our MI regularization is concise and efficient.

This work extends our previous work \cite{liu2021identity} in the following significant ways:

\textbf{$\bullet$} A novel expression and identity disentanglement framework based on practical mutual information minimization, which follows a min-min game with the joint and marginal distribution sampling.

\textbf{$\bullet$} We demonstrate the generality of our framework in more FER dataset, i.e., MMI, carry out all experiments using the novel MIC framework. 

\textbf{$\bullet$} Moreover, the systematical cross-dataset evaluation, sensitivity analysis, and identity feature extraction analysis are provided.

The main contributions of this paper are summarized as follows:

\textbf{$\bullet$} We propose to inference expression from the residual frames, which explores the off-the-shelf yet valuable expression related muscle movement in the up to two orders of magnitude compressed domain.

\textbf{$\bullet$} Targeting for the identity-aware video-based FER, the independence of expression and identity representations from P frames and I frame are enforced with the differentiable MI measure.

\textbf{$\bullet$} The separability of expression and identity representations is maximized by a min-min game with the joint and marginal distribution sampling, which does not rely on the unstable adversarial game to achieve identity elimination.

We evidenced its effectiveness on several video-based FER benchmarks with much faster inference. The promising performance evidenced its generality and scalability.


\section{Related Works}

\noindent{\bf{Video-based FER}} has been thoroughly researched, since facial expression is a natural and universal means for human communication \citep{li2020deep,Jiyoung2020multi}. Considering that the expression is essentially a dynamic action which should take minute muscle movements through time into account \citep{sandbach2012static}. Traditionally, the handcrafted features are utilized to represent the spatio-temporal cues and for FER. Frame aggregation and spatiotemporal FER networks are being developed in parallel with the exponential growth of deep learning. Frame aggregation approaches may make use of image-based FER networks by conducting frame-wise aggregation at the decision-level \citep{kahou2016emonets} or feature-level \citep{xu2016video}.

The essential temporal correlation, on the other hand, is not investigated. Instead, the spatio-temporal FER networks use sequential frames to utilize both the spatial/textural and temporal information \citep{al2018deep}. Furthermore, cascaded networks suggest integrating CNN-learned perceptual vision representations with RNNs for variable-length video \citep{donahue2015long,jain2017multi}. Moreover, by using a non-local network for video processing, the number of potential connections and the corresponding computation costs grow exponentially to the number of frames. The 3D convolutional has shared kernel-weights along the time axis, which has also been widely used for video-based FER \citep{barros2016developing,zhao2018learning}.

However, recent studies have only looked at the image domain, and the spatiotemporal FER does not outperform aggregation methods substantially \citep{li2020deep}. To the best of our knowledge, this is the first effort to investigate the compressed video FER, which is orthogonal to these advantages and can be conveniently added to each other.

\begin{figure}[t]
\centering
\includegraphics[width=8cm]{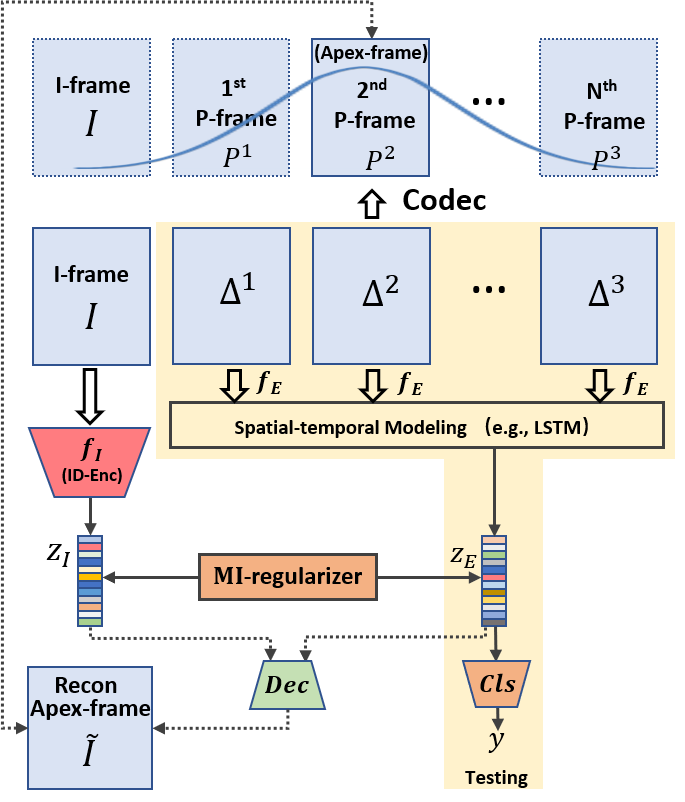}\\
\caption{The illustration of our \textbf{M}utual \textbf{I}nformation Regularized in \textbf{C}ompressed Video (MIC) framework for identity-aware FER in the compressed video domain. Dec and Cls indicate decoder and classifier, respectively.}\label{fig:2}
\end{figure}

\noindent{\bf{Video compression}} convert the digital video into a specific format that is suitable for recording and distribution of this video \citep{richardson2003h264}. Conventionally, the H.264/MPEG-4 and the Advanced Video Coding are the typical algorithms \citep{yeo2006compressed}. Typically, the video sequence is divided into several Group Of Pictures (GOP) by the video codecs. In a GOP, there is an I frame and follow by several P frames. Specifically, the I frame is a self-contained RGB frame with full visual representation, while the P frame is the inter frames that hold motion vectors and residuals w.r.t. the previous frame \citep{le1991mpeg}. The motion vector can be used as an alternative to the optical flow \citep{zhang2018real,zhang2016real}, which needs to decode the RGB images.

The recent action recognition method \citep{wu2018compressed} proposes to aggregate I frame, residuals, and motion vectors in the compressed domain, without the RGB image decoding. Although FER shares some similarities with action recognition, the movement range and use of I frame can be vastly different. In action recognition \citep{wu2018compressed}, the I frame is directly used to predict the action and combine with the result of P frames. In contrast, the I frame in FER usually be a neutral face (different from the video label). Furthermore, the low-resolution motion vector can not well encode the expression.     


\noindent{\bf{Mutual-information}} has a long history in unsupervised learning. The infomax principle~\citep{linsker1988self}, as prescribed for neural networks, advocates maximizing the MI between network input and output. This can be the fundamental of many ICA algorithms, which can be nonlinear~\citep{hyvarinen1999nonlinear} but are often hard to adapt for use with deep networks. Recently, some of the works proposed to achieve unsupervised learning with MI. \citep{brakel2017learning} proposes a generative adversarial network to minimize MI with positive and negative samples for Independent Component Analysis (ICA). It introduces a strategy to draw samples from the joint distribution and the product of marginal distributions and proposed to train an encoder and a discriminator to minimize the Jansen-Shannon divergence. Moreover, it has recently been shown that the GAN framework can be extended not only to maximize or minimize MI but also to explicitly compute it using the Mutual Information Neural Estimation (MINE) proposed in \citep{belghazi2018mine}. In \citep{hjelm2018learning}, the DeepInfoMax (DIM) is proposed to learn the representations based on both local and global information. In \citep{velivckovic2018deep}, Deep Graph Infomax (DGI) extends this approach to graph-structured data.

Inspired by these works, we are targeting to utilize the MI as the justified measure of independence, and minimize it directly as our disentanglement objective.


\noindent{\bf{Eliminating identity}} can benefit to extract more ``pure" expression feature \citep{meng2017identity,liu2017adaptive}. We note that previous identity-aware FER methods usually explicitly require the identity labels of FER datasets to sample the triplets \citep{meng2017identity,liu2017adaptive,liu2018adaptive}, while the identity label is not common in FER. In contrast, our solution does not relies on the identity label of FER samples, but utilizes the easily available face recognition dataset.

The typical solution for FER is metric learning. Our MI regularization is also related to the triplet loss \citep{schroff2015facenet}, which maximizes the Euclidean or cosine distance between two identities. With the development of GAN, adversarial training also can be utilized for disentanglement \citep{liu2019feature,liu2021mutual}. Instead, we consider the mutual information to be a more meaningful divergence to capture complex non-linear relationships, between the identity and expression representations. Besides, the identity label is required in these methods \citep{meng2017identity,liu2017adaptive,liu2018adaptive}. We can also choose adversarial training \citep{liu2019feature,liu2021mutual} as a baseline to achieve identity elimination in our framework. We note that the adversarial game is notorious for its unstable model collapse \citep{goodfellow2016nips}, while our solution follows a collaborative way.

Several adversarial disentanglement works demonstrate that simply separate the input may result in the extracted feature has no meaningful information \citep{liu2021mutual,liu2019feature,liu2018exploring,hadad2018two}. The reconstruction of input can explicitly enforce the disentangled factors to be complementary to each other. However, reconstructing the video can be hugely underconstrained.

\section{Methodology}

We propose to develop an efficient video-based FER framework that operates directly on the compressed domain. The overall framework is shown in Figure. 2, which is consisted of four core modules. The pre-trained identity branch and FER branch (frame embedding network $f_E$, aggregation module, and Classifier) work on the I frame and undecoded P frames, respectively. The dependence of identity and expression representations is then measured using a differentiable mutual information regularization module. Furthermore, when the apex frame is annotated, the complementary constraint can be applied to stabilize the early stage training.

\subsection{Modeling Compressed Representations}

To illustrate the format of the input video, we choose the MPEG-4 as an example \citep{richardson2003h264}. The compressed domain has two typical frames, i.e., I frame and P frames. Specifically, the I frame $I\in\mathbb{R}^{h\times w\times 3}$ is a complete RGB image. We use $h$ and $w$ to denote its height and width, respectively. Besides, the P frame at time $t$ $P^t\in\mathbb{R}^{h\times w\times 3}$ can be reconstructed with the stored offsets, called residual errors $\Delta ^t\in\mathbb{R}^{h\times w\times 3}$ and motion vectors $\mathcal{T}^t\in\mathbb{R}^{h\times w\times 2}$.

Noticing that the motion vectors $\mathcal{T}^t$ has a much lower resolution, since their values within the same macroblock are identical. Considering the micro-movements of facial expression in each frame, the low resolution $\mathcal{T}^t$ usually not helpful for the FER. 
For P frame reconstruction $P_i^t=P_{i-\mathcal{T}_i^t}^{t-1}+\Delta_i^t$, where index all the pixels and $P^0=I$. Then, $\mathcal{T}^t$ and $\Delta ^t$ are processed by discrete cosine transform and entropy-encoded.


The majority of compression algorithms are programmed primarily to minimize file size, and the encoded format can vary greatly from RGB images in terms of statistical and structural properties. As a consequence, a specially built processing network is needed to manage the compressed format. Considering the structure of residual images $\Delta ^t$ are much simpler than the decoded images, it is possible to utilize simpler and faster CNNs $f_E:\mathbb{R}^{h\times w\times 3}\rightarrow \mathbb{R}^{512}$ to extract the feature of each frame \citep{kim2017multi,baddar2019mode,baddar2019fly}. Practically, we follow the 
CNN in the typical CNN-LSTM FER structure \citep{baddar2019mode,kim2017multi,baddar2019fly}, but with fewer layers to explore the information in $\Delta^t$. Noticing that $f_E$ is shared for all frames, and only needs to store one $f_E$ in processing.

Besides, most existing action recognition methods with compressed video \citep{wu2018compressed} independently concatenate the paired $\Delta ^t$ and $\mathcal{T}^t$ at each time step and predict an action score of each P-frame. The temporal cues and their development patterns are important for the FER task \citep{li2020deep}. We simply choose the LSTM in \citep{baddar2019mode} to model the sequential development of residual frames and summarize the information to an expression feature $z_E$. Since our LSTM is applied to 512-dim features, the computation burden is largely smaller than work on the raw images. Noticing that more advanced RNN, 3D CNN, or attention networks can potentially be utilized to replace our LSTM model to further boost the performance \citep{barros2016developing,zhao2018learning,kumawat2019lbvcnn}.

For the I frame with raw image format, we simply use the FaceNet \citep{schroff2015facenet} pre-trained on millions of identities \citep{kemelmacher2016megaface} as our identity feature extractor $f_{I}:\mathbb{R}^{h\times w\times 3}\rightarrow z_{I},$ where $z_{I}\in\mathbb{R}^{1024}$ denotes the identity feature. We note that $Z_E$ and $z_I$ do not need to have the same dimension for MI Regularization. Several datasets' FER videos begin with a neutral expression, which may help with identity recognition.

\begin{algorithm}[t!]
\caption{Training scheme of our framework}\label{alg:A2}
\begin{algorithmic}[1]
\State Initialize network parameters of $f_E$, $LSTM$, $cls$, $T_\theta$ and $Dec$.
\While {Not Converged}
        \State Randomly sample $n$ compressed FE videos.
        \State Extract expression and identity feature $z_E$ and $z_I$.
        \State Draw $n$ $(z_E,z_I)$ for the joint distribution.
        \State Draw $n$ $z_I$ for the marginal distribution.
        \State Evaluate lower bound $MI{\widehat{(z_{E};z_{I})}}_n$
        \State $=\frac{1}{n} \sum_{i=1}^{n} T(z_{E},z_{I},\theta)-{\rm{log}}(\frac{1}{n} \sum_{i=1}^{n} e^{T(z_{E},\hat{z_{I}},\theta)})$.
        \State Calculate the cross entropy loss for $n$ samples
        \State$\mathcal{L}_{CE} = \frac{1}{n} \sum_{i=1}^{n}\{-\sum_{c=1}^C y_{c}\log(Cls(z_E)_{c})\}$.
        \State Reconstruct $\hat{I}_{Apex}$ using $z_E$ and $z_I$.
        \State Calculate the cross entropy loss for $n$ samples 
        \State$\mathcal{L}_1 = \frac{1}{n} \sum_{i=1}^{n}||I_{Apex}-\hat{I}_{Apex}||_2^2$.

        \State Compute gradients and update network parameters.
        \State Cls $\leftarrow \nabla\mathcal{L}_{CE}$ 
        \State $f_E$ and LSTM $\leftarrow \nabla\mathcal{L}_{CE}+\nabla  \alpha MI{\widehat{(z_{E};z_{I})}}_n+ \nabla\beta \mathcal{L}_1$. 
        \State $T_\theta \leftarrow \nabla MI{\widehat{(z_{E};z_{I})}}_n$. 
        \State $Dec \leftarrow \nabla \mathcal{L}_1$.

\EndWhile
\end{algorithmic}
\label{alg:alg1}
\end{algorithm}

\subsection{MI regularization}

To eliminate the identity-related factors in our FER representation, we propose to utilize the identity feature from pre-trained face recognizer $z_I$ as anchor, and explicitly inspect the information w.r.t. $z_I$ in $z_E$.  

Achieving the disentanglement of different factors requires two major objectives, i.e., 1) each factor has its specific information, and 2) does not incorporate the information of the other factors \citep{liu2019feature,liu2021mutual}. For example, $z_{E}$ can achieve 1) using the conventional C.E. loss minimization w.r.t. the expression label $y$, and $z_{I}$ is from the pre-trained identity extractor, which inherently has identity information. However, how to explicitly measure the dependency between these factors and minimize this metric to achieve the latter objective can be challenging.

Actually, the mutual information (MI) is the exact metric to measure the amount of information obtained about one random variable through observing another random variable. \begin{align}
MI(z_{E};z_{I})= \int_{\mathcal{E}\times\mathcal{I}} {\rm{log}} \frac{d\mathbb{P}_{z_{E}z_{I}}}{d\mathbb{P}_{z_{E}}\otimes\mathbb{P}_{z_{I}}}{d\mathbb{P}_{z_{E}z_{I}}}
\end{align} where $z_{E}$ and $z_{I}$ are the random variables follows the distribution $\mathcal{E}$ and $\mathcal{I}$ respectively. ${\mathbb{P}_{z_{E}z_{I}}}$ indicates the joint probability distribution of $(z_{E},z_{I})$, $\mathbb{P}_{z_{E}}=\int_\mathcal{I} {d\mathbb{P}_{z_{E}z_{I}}}$ and $\mathbb{P}_{z_{I}}=\int_\mathcal{E} {d\mathbb{P}_{z_{E}z_{I}}}$ are the marginals. ${\mathbb{P}_{z_{E}}\otimes\mathbb{P}_{z_{I}}}$ is the product of the marginals. 

MI minimization explicitly enforces the joint distribution to be equal to the product of marginals, which leads to the statistical independence of two vectors. Instead, the MI maximization can result in two vectors have the same information, and the MI is simply equal to the entropy of a variable.

\begin{figure*}[t]
\centering
\includegraphics[width=16cm]{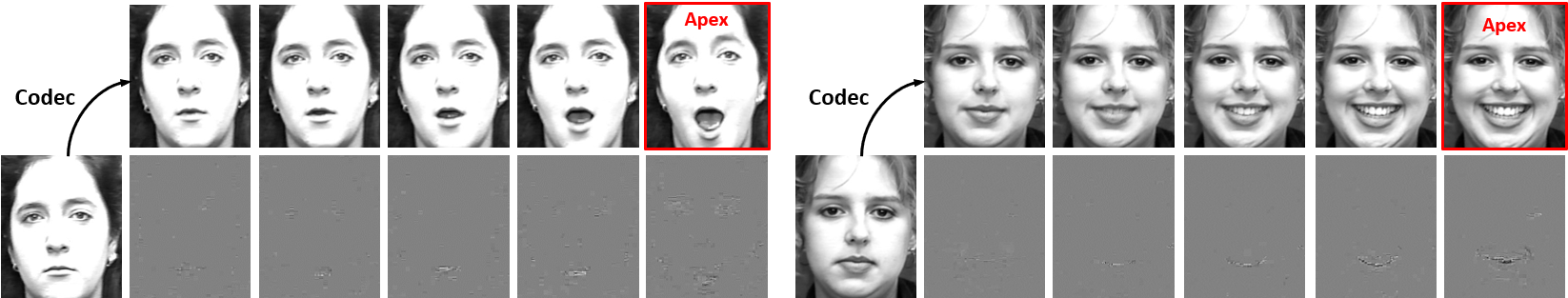}\\
\caption{The illustration of the compressed and decoded frames in CK+ dataset.}\label{fig:3} 
\end{figure*}
 
We propose to utilize the mutual information neural estimator (MINE) \citep{belghazi2018mine} to provide the unbiased estimation of MI on $n$ independent and identically distributed (i.i.d.) samples. It is linearly scalable w.r.t. dimensionality and sample size, by leveraging a gradient descent over neural network $T_\theta:\mathcal{E}\times\mathcal{I}\rightarrow \mathbb{R}$. MINE proposes to approximate MI by exploiting a lower bound based on the \textit{Donsker-Varadhan} representation of the Kullback-Leibler divergence. Therefore, the neural information measure can be formulated as \begin{eqnarray}
MI{\widehat{(z_{E};z_{I})}}_n= {\mathop{}^{\rm{sup}}_{\theta\in{\Theta}}} \left\{\mathbb{E}_{\mathbb{P}^{n}_{z_{E}z_{I}}}[T_\theta]-{\rm{log}}(\mathbb{E}_{\mathbb{P}^{n}_{z_{E}}\otimes{\hat{\mathbb{P}}^{n}_{z_{E}}}}[e^{T_\theta}])\right\}
\end{eqnarray} Given the distribution $\mathbb{P}$, ${\hat{\mathbb{P}}^{n}_{z_{E}}}$ denotes the empirical distribution associated to $n$ i.i.d. samples. Since the supremum is taken over all functions of $T$, the two expectations are finite. Then, the MI in Eq. (2) can be estimated as follows:
\begin{eqnarray}
MI{\widehat{(z_{E};z_{I})}}_n =\int\int\mathbb{P}^{n}_{z_{E}z_{I}}(z_{E},z_{I})T(z_{E},z_{I},\theta)\nonumber\\
-{\rm{log}}(\int\int\mathbb{P}^{n}_{z_{E}}(z_{E})\mathbb{P}^{n}_{z_{I}}(z_{I})[e^{T(z_{E},z_{I},\theta)]})
\end{eqnarray}

Besides, we leverage Monte-Carlo integration to avoid computing the integrals to compute $MI{\widehat{(z_{E};z_{I})}}_n$ as
\begin{eqnarray}
\frac{1}{n} \sum_{i=1}^{n} T(z_{E},z_{I},\theta)-{\rm{log}}(\frac{1}{n} \sum_{i=1}^{n} e^{T(z_{E},\hat{z_{I}},\theta)})
\end{eqnarray} where $\hat{z_{I}}$ is sampled from the marginal distribution. Note that $(z_{E},z_{I})$ are sampled from the joint distribution ${\mathbb{P}_{z_{E}z_{I}}}$. Then, we can evaluate the bias corrected gradients (e.g., moving average) \citep{belghazi2018mine}. 

The estimated $MI(z_{E};z_{I})$ is used as the supervision to update the FER branch. By utilizing MI regularization, the adversarial discriminator \citep{xie2017adversarial,liu2019feature,liu2021mutual} is no longer needed in our new framework, which makes the balance of each module easier. Note that we need the additional neural network $T_\theta$ to measure the MI, but it is collaboratively trained with the FER branch to maximize the discrepancy between the two features. Essentially, we are playing a \textit{min-min} game instead of a \textit{min-max} game. Therefore, it is easier to stabilize the training (compared to adversarial training).

Besides, MI is a symmetric measure, while the conditional entropy $H(z_{I}|z_{E})=H(z_{I})-MI(z_{I};z_{E})$ optimized in conventional disentanglement works \citep{liu2019feature,liu2021mutual} is asymmetric and essentially we should calculate both $H(z_{I}|z_{E})$ and $H(z_{I}|z_{E})$ as supervision signal \citep{liu2019feature,liu2021mutual}.


To maximize the discrepancy of $z_{E}$ and $z_{I}$, we can also apply the adversarial disentanglement solutions \citep{liu2019feature,liu2021mutual}. Nevertheless, with the above-mentioned limitations, such methods can be hard to optimize and lead to inferior performance.

\subsection{Complementary constraint}

Many FER datasets follow a well-defined collection protocol, which usually starts from the neutral face and then develops to an expression. Specifically, the video in CK+ \citep{kanade2000comprehensive,lucey2010extended} consists of a sequence that shifts from the neutral expression to an apex facial expression. The last frame usually is the apex frame, which has the most strong expression intensity. Actually, the image-based FER methods select the last three frames to construct their training and testing datasets. Similarly, in MMI \citep{pantic2005web}, the video frames usually start from the neutral face and develop to the apex around the middle of the video, and returning back to the neutral at the end of the video. Noticing that the apex frame (i.e., last frame in CK+ or middle frame in MMI) can clearly incorporate both the identity and expression information. Therefore, we are possible to utilize the apex frame as a reference of reconstruction, and simply apply the $\mathcal{L}_2$ loss.  \begin{eqnarray} \label{eq:l1}
\mathcal{L}_1 = ||I_{Apex}-\hat{I}_{Apex}||_2^2
\end{eqnarray}where $\hat{I}_{Apex}=Dec(z_{I},z_{E})$. The complementary restriction is not necessary for our system since the FER loss is heavily weighted in the FER branch. It requires to maintain sufficient information w.r.t. expression and not easy to have nothing meaningful. However, the complementary constraint does helpful for the convergence in the early stage. When the apex is annotated, we only need to decode the apex frame in the decoded image domain at the start of a few training epochs.

\subsection{Overall objectives}

We have three to be minimized objectives, i.e., cross-entropy loss, mutual information and $\mathcal{L}_1$ loss, which works collaboratively to update each module. The expression classification is the main task of the FER. We choose the typical cross-entropy loss $\mathcal{L}_{CE} = -\sum_{c=1}^C y_{c}\log(Cls(z_E)_{c})$ to ensure $z_E$ contains sufficient expression information and finally have a good performance on $C$-class expression classification. We use $y_{c}$ and $Cls(z_E)_{c}$indicate the $c^{th}$ class probability of the label and classifier softmax predictions respectively. Since the FER branch can be updated with all of the losses, we assign the balance parameter $\alpha\in[0,1]$ and $\beta\in[0,1]$ to mutual information and $\mathcal{L}_1$ loss minimization objectives, respectively. Specifically, we update our $f_E$ and LSTM modules with 
\begin{align}
     \mathcal{L}_{CE}+ \alpha MI{\widehat{(z_{E};z_{I})}}_n+ \beta \mathcal{L}_1
\end{align} For the MI estimator $T_\theta$, we update it with $ MI{\widehat{(z_{E};z_{I})}}_n$. Moreover, the decoder module $Dec$ is updated with $\mathcal{L}_1$. The detailed training flow is shown in Algorithm \ref{alg:A2}. We note that only the FER branch, i.e., $f_E$, LSTM and Cls, is used for testing.

Since we are using the neural network $T_\theta$ for mutual information estimation, it is scalable, flexible, and completely trainable via back-propagation. Moreover, the decoder for reconstruction is used for complementary constraints. In contrast, FLF \cite{liu2019feature,liu2021mutual} uses a discriminator and a decoder for adversarial disentanglement and complementary constraint. Considering the $T_\theta$ in MIC and discriminator in FLF \cite{liu2019feature,liu2021mutual} has a similar structure, there is no significant difference for the network complexity. The mutual information calculation with Eq. 4 has the complexity of $\mathcal{O}(n)$, where $n$ is the number of sampled data. The linear complexity can be simple for implementation. Conventionally, the fast computation of MI is limited to discrete variables \citep{paninski2003estimation}. For the continuous random variables, its complexity is quadratic to the number of samples, which is not desired for a loss function.

\begin{figure*}[t]
\centering
\includegraphics[width=16cm]{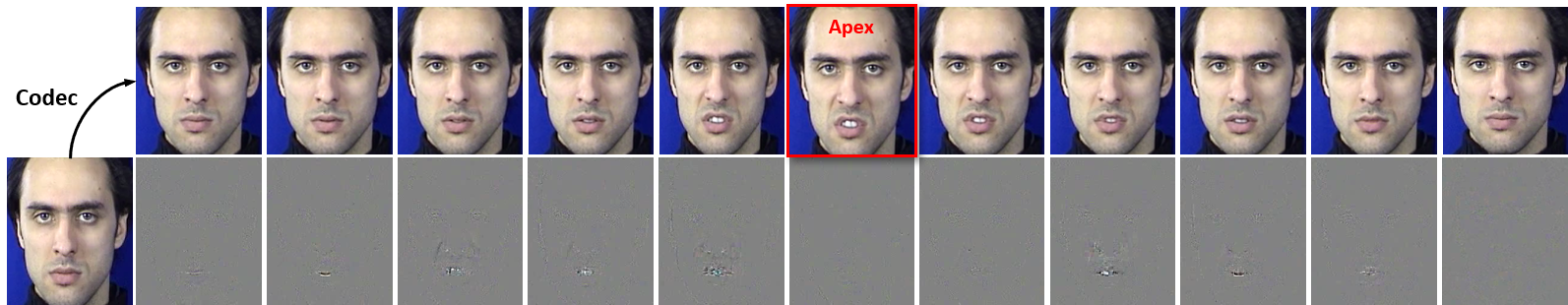}\\
\caption{The illustration of the compressed and decoded frames in MMI dataset.}\label{fig:4} 
\end{figure*}

\section{Experiments}\label{sect:exp}

In this section, we first detail our experimental setup, present a quantitative analysis of our model, and finally compare it with state-of-the-art methods. The good FER accuracy and high inference speed in testing demonstrate its effectiveness.

\begin{table}[t]
	\centering
	\scriptsize
	\begin{tabular}{| l | c | c | c |}
		\hline
		\textbf{Method} & \textbf{Accuracy}  & \textbf{Landmarks} & \textbf{Ave Test} \\
		\hline \hline

		PHRNN-MSCNN (2017) \citep{zhang2017facial}   & 98.50 & $\checkmark$   & -\\
		C3D-GRU (2019) \citep{lee2019visual} & 97.25 &$\times$		 & - \\
		CTSLSTM (2019) \citep{hu2019video}	& 93.9 & $\checkmark$       & - \\ 			
		(N+M)-tuplet (2019) \citep{liu2018adaptive}   & 93.90 & $\checkmark$   & 12fps  \\
		SC (2019) \citep{verma2019facial}		& 97.60 & $\checkmark$      & - \\ 	
		G2-VER (2019) \citep{albrici2019g2}			& 97.40 & $\times$       & - \\ 			
		LBVCNN (2019) \citep{kumawat2019lbvcnn}		& 97.38 & $\times$       & - \\ 
		NST (2020) \citep{kumar2020noisy}$\dag$		& \textbf{99.69} & $\times$       & - \\  		
		\hline		\hline 

Mode VLSTM (2019) \citep{baddar2019mode} & 97.42 & $\times$       & 11fps   \\\hline
MIC  & \textbf{98.95} & $\times$       & \textbf{35fps}     \\
MIC-MI & 97.84 & $\times$       & \textbf{35fps}    \\

MIC-MI+Adv\citep{liu2019feature}   & 98.78 & $\times$       & \textbf{35fps}  \\
MIC-$\hat{I}$  & 98.72 & $\times$       & \textbf{35fps}   \\
MIC+$\mathcal{T}^t$  & 98.93 & $\times$       & 29fps     \\\hline	
		
FAN+ResNet18* (2019) \citep{meng2019frame}& 99.69 &$\times$&   {10fps}\\
MIC+ResNet6 & \textbf{99.71} &$\times$&  \textbf{31fps}   \\ \hline	
		
	\end{tabular}
	\vspace{0.1in}
	\caption{Experimental results on the CK+ dataset. Note that in order to make the comparison fair, we do not consider image-based and 3D geometry based experiment setting and models \citep{liu2017adaptive,meng2017identity,liu2019hard}.*Additional FER+ dataset is used. $\dag$Additional body language dataset is used.}
	\label{tab:ck_all}
\end{table}

\subsection{Description of the datasets}

\noindent\textbf{CK+ Dataset \citep{kanade2000comprehensive,lucey2010extended}} is referring to the Cohn-Kanade AU-Coded Expression dataset, which is a widely accepted FER benchmark \citep{li2020deep,kumawat2019lbvcnn}. The video is collected in a restricted environment, in which the participate subjects are facing the recorder with an empty background. The video in CK+ consists of a sequence that shifts from the neutral expression to an apex facial expression. The last frame usually is the apex frame, which has the most strong expression intensity. The expression included in this dataset is anger, contempt, disgust, fear, happiness, sadness, and surprise. There are 327 facial expression videos collected from 118 subjects. Following the previous works, we use subject independent 10-folds cross-validation \citep{li2020deep,liu2018adaptive}.

Many FER datasets follow a well-defined collection protocol, which usually starts from the neutral face and develops to an expression. Specifically, the image-based FER methods select the last three frames to construct their training and testing datasets. In Fig. \ref{fig:3}, we show the compressed and decoded frames in CK+ dataset. Similarly, in MMI \citep{pantic2005web}, the video frames usually start from the neutral face and develop to the apex around the middle of the video, and returning back to the neutral at the end of the video. Noticing that the apex frame (i.e., last frame in CK+ or middle frame in MMI) can clearly incorporate both the identity and expression information. Therefore, we are possible to utilize the apex frame as a reference of reconstruction, and simply apply the $\mathcal{L}_1$ loss.

\noindent\textbf{MMI Dataset \citep{pantic2005web}} consist of a total of 326 facial expression videos from 32 participants. There are 213 labeled videos with the expression label angry, disgust, fear, happy, sad, and surprise. The video frames start from the neutral face. Then the expression is developed to the apex in the middle of video, and returning back to the neutral at the end of the video. In our experiments, we follow the previous works to use subject independent 10-folds cross-validation \citep{li2020deep,liu2018adaptive}. In Fig. \ref{fig:4}, we show the compressed and decoded frames in the MMI dataset.

\begin{figure*}[t]
\centering
\includegraphics[width=16cm]{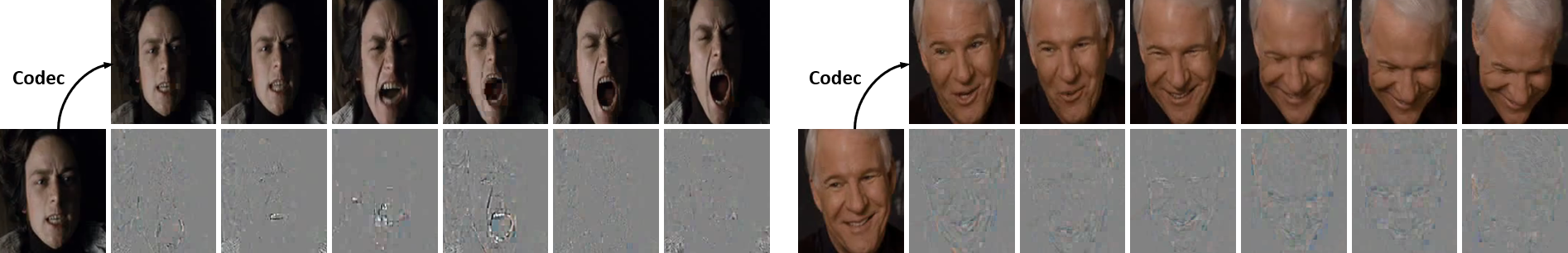}\\
\caption{The illustration of the compressed and decoded frames in AFEW dataset.}\label{fig:5} 
\end{figure*}

\noindent\textbf{AFEW Dataset \citep{dhall2014emotion}} is more close to the uncontrolled real-world environment. It is consists of video clips of movies \citep{perveen2018spontaneous}. The video in AFEW has a spontaneous facial expression. The AFEW has seven expressions: anger, disgust, fear, happiness, sadness, surprise, and neutral. Following the evaluation protocol in EmotiW \citep{dhall2017individual}, there are training, validation, and testing sets. Since its testing label is not available, we follow the previous work to use the validation set for comparison \citep{baddar2019mode}. Noticing that the validation set is not used in the training stage for the parameter or hyper-parameter tuning. In Fig. \ref{fig:5}, we show the compressed and decoded frames in the AFEW dataset.

\begin{table}[t]
	\centering
	\footnotesize
	\begin{tabular}{| l | c | c |}
		\hline
		\textbf{Method} & \textbf{ID accuracy}  & \textbf{MI}  \\
		\hline \hline
Mode variational LSTM (2019) \citep{baddar2019mode} & 28.4 & 1.57    \\\hline
MIC  & \textbf{0.8} &     \textbf{0.01}   \\
MIC-MI & 5.7 & 1.22         \\

MIC-MI+Adv\citep{liu2019feature}   & 1.4   & 0.08       \\
MIC-$\hat{I}$  & 0.9  & 0.10        \\
MIC+$\mathcal{T}^t$  & \textbf{0.8}  & 0.09     \\\hline	

	\end{tabular}
	\vspace{0.1in}
	\caption{Comparison of the identity eliminating on CK+ dataset w.r.t. the identity recognition accuracy using $z_E$, and the mutual information between $z_E$ and $z_I$.}
	\label{tab:idck}
\end{table}

\subsection{Implementation details} 

We preprocess video frames and augment the data according to  \citep{baddar2019mode,kim2017multi,baddar2019fly} for fair comparison. For these three datasets, the videos only have one GOP and do not need to segment the video. We utilize the Pytorch deep learning platform for our framework. In the training stage, on all datasets, we set the batch size to 48. All of the modules use the Adam optimizer with momentum 0.9, and a weight decay of 1e-5 for 100 training epochs. On the CK+ and MMI datasets, the learning rate is initialized to 1e-1, and be modified to 1e-2 for the 30$^{th}$ epoch. For the AFEW dataset, we initialize the learning rate to 1e-4, and modify it to 8e-6 for the 30$^{th}$ epoch and 1e-7 for the 60$^{th}$ epochs.

All of our training/testings use an NVIDIA Titan X GPU. We note that the calculation of the accumulated residuals to recover the apex frame is measured on Intel E5-2698 v4 CPUs, but we do not need this operation in testing. For the testing speed, we measure the average frame per second (fps) according to the average running time, which is the sum of the data pre-processing time and the FER branch forward pass time.

Practically, our $f_E$, LSTM and Cls follow the CNN-LSTM structure in \citep{kim2017multi,baddar2019mode,baddar2019fly} for fair comparison. Considering the relatively simpler residual data, we use part of convolutional layer for $f_E$ (i.e., C1, C2 and F4 layers as in \citep{kim2017multi,baddar2019mode,baddar2019fly}) and a fully connected layer for Cls (i.e., $\mathbb{R}^{512}\rightarrow\mathbb{R}^6 or \mathbb{R}^7$).

FAN \cite{meng2019frame} shows the ResNet \cite{he2016deep} can be a powerful backbone. To further demonstrate the generality of our framework, we use the first five convolutional layers (i.e., before the second residual block) and the first fully connected layer in ResNet18 as our feature extractor backbone. We denote this ResNet backbone as ResNet6.

For the CK+ and MMI dataset, we choose the last or middle frame as the apex reference image, respectively. Since the complementary constraint is only used to stabilize the initial training of disentanglement, we uniformly decrease $\beta$ from 1 to 0 until the $30^{th}$ epoch. Practically, we use the grid search to find the optimal $\alpha$ and set it to 0.1, 0.1, 0.2 on CK+, MMI, and AFEW datasets, respectively.

\subsection{Evaluation and ablation study}

\noindent\textbf{Results on CK+ dataset.} 

The 10-fold cross-validation performance of our proposed method is shown in Table 1. For a fair comparison, the image-based experiment settings are not incorporated in the tables. Besides, only the state-of-the-art (SOTA) accuracy obtained by the single-models (non-ensemble model) is listed.

Many models, e.g., PHRNN-MSCNN \citep{zhang2017facial}, CTSLSTM \citep{hu2019video} and SC \citep{verma2019facial}, achieved the SOTA performance by utilizing the facial landmarks. However, this operation highly relies on fine-grained landmark detection, which itself is a challenging task \citep{li2020deep,liu2019hard}, and unavoidably introduced additional computation.  

Based on the mode variational LSTM \citep{baddar2019mode}, our proposed MIC achieves the SOTA result without the facial landmarks, 3D face models, or optical flow. It worth noticing that the much simpler CNN encoding network makes more residual frames that can be processed parallel than \citep{baddar2019mode}. Moreover, our efficiency also benefits from avoiding the decompress of the video. Since the videos are typically stored and transmitted with the compressed version, and the residuals are off-the-shelf. As a result, the proposed compressed domain MIC can speed up the testing about 3 times over \citep{baddar2019mode}, and achieve better accuracy.

\begin{table}[t]
	\centering
	\scriptsize
	\begin{tabular}{| l | c | c | c |}
		\hline
		\textbf{Method} & \textbf{Accuracy}  & \textbf{Landmarks} & \textbf{Ave Test} \\
		\hline \hline

3D CNN-DAP (2014) \citep{liu2014deeply} & 63.4& $\checkmark$ &  \\
CNN+LSTM (2017) \citep{kim2017multi}   & 78.61&  $\times$ &  \\
CTSLSTM (2019) \citep{hu2019video}   &78.40& $\checkmark$& 8fps \\\hline

Mode VLSTM (2019) \citep{baddar2019mode}&79.33 &  $\times$ & 10fps  \\\hline
MIC  & \textbf{81.29} &  $\times$ &   \textbf{32fps}  \\
MIC-MI  & 80.25 &  $\times$ & \textbf{32fps} \\

MIC-MI+Adv\citep{liu2019feature} & 80.98 &  $\times$ &   \textbf{32fps}  \\
MIC-$\hat{I}$  & 80.94&  $\times$ &  \textbf{32fps} \\
MIC+$\mathcal{T}^t$  & 81.24&  $\times$ &  28fps  \\\hline
		
	\end{tabular}
	\vspace{0.1in}
	\caption{Experimental results on MMI dataset. Note that in order to make the comparison fair, we do not consider image-based and 3D geometry based experiment setting and models \citep{liu2017adaptive,meng2017identity,liu2019hard}.}
	\label{tab:ck_all}
\end{table}

Besides, \citep{liu2018adaptive} is a typical metric-learning-based identity removing method. Our solution can significantly outperform it with respect to both speed and accuracy. Actually, the sampling of tuplets usually makes the training not scalable \citep{liu2019hard}, while our identity eliminating scheme is concise and effective.

When we remove some modules from our framework, the performances have different degrees of decline. We use -MI and -$\hat{I}$ to denote the MIC without MI regularizer or complementary constraint, respectively. The performance drop of MIC is significant when we remove the MI regularization module, which further evidenced that the identity can be a notorious factor for FER. The result also implies that the identity can be well encoded by the face recognition network and the disentanglement with mutual information is feasible. Compared with using the conventional adversarial training based disentanglement \citep{liu2019feature,liu2021mutual} as an alternative (i.e., MIC-MI+Adv), our MI regularizer is easy to train and can converge 1.8 times faster in training.

We can also follow the action recognition method \citep{shou2019dmc} to concatenate the motion vector and residual as the input, and denote as MIC+$\mathcal{T}^t$. However, we do not achieve significant performance on all datasets, but the inference speed in testing can be slower. This may be related to the coarse resolution of the motion vector can not well describe the fine-grain muscle movement of the face. More appealingly, the performance of our MIC can be further improved With the ResNet backbone. With the simplified 6-layer ResNet, MIC outperforms the FAN \cite{meng2019frame} w.r.t. both accuracy and processing speed. Aggregation-based methods do not explore the temporal cues and can be computational costly to compare all possible image pairs within a set \cite{liu2019permutation}. We note that the ResNet18 used in FAN \cite{meng2019frame} is sequentially pre-trained on the additional MSCeleb-1M face recognition dataset and FER+ expression dataset.

The confusion matrix of our proposed MIC method on the CK+ is reported in Figure 7 (left). The accuracy for the expression of class happiness, disgust, anger, surprise, and contempt are almost perfect.

\begin{figure}[t]
	\centering
	\includegraphics[width=1\columnwidth]{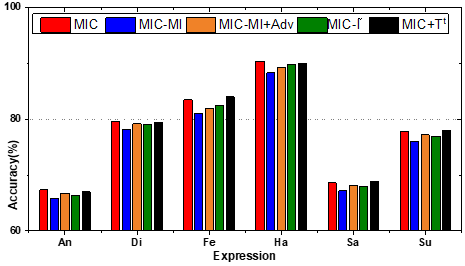}
	\caption{Experimental results on MMI dataset accuracy according to each emotion among five networks.}
	\label{fig:oulu}
\end{figure}

In Table \ref{tab:idck}, we investigate the identity eliminating performance. The first metric is following \citep{liu2019feature} to use recognize identity with $z_E$. Besides, we can directly use mutual information as the metric of independence. We can see that the residual frame itself can incorporate much less identity information than the decoded images as in \citep{baddar2019mode}, while it is still possible to detect identity with the facial contours. The MI regularization can explicitly remove the identity factors and outperforms the adversarial training \citep{liu2019feature}.

\begin{figure*}[t]
\centering
\includegraphics[width=17cm]{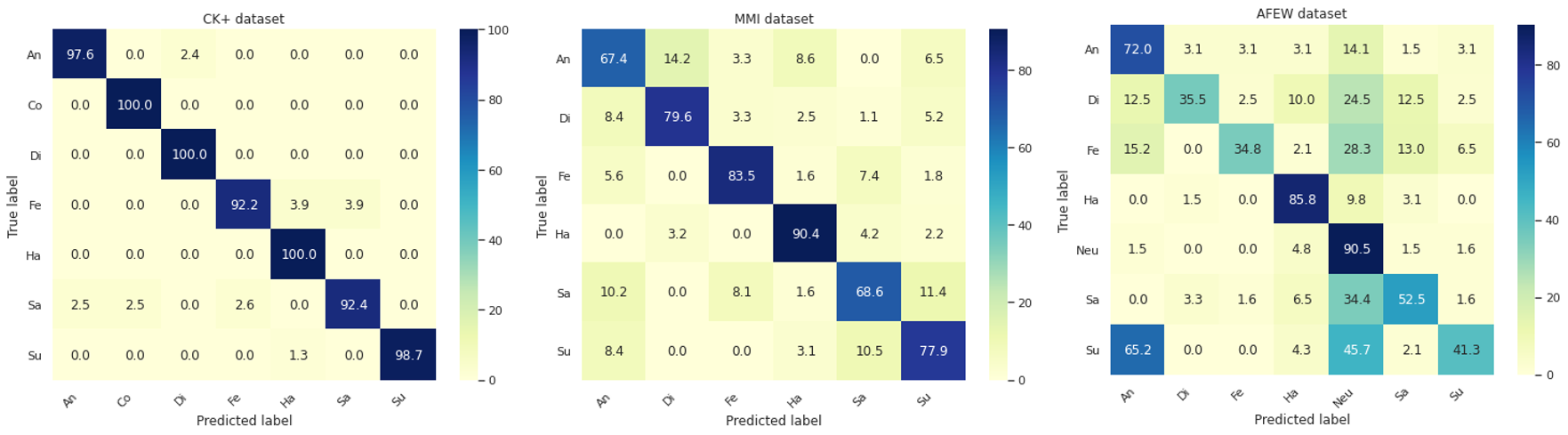}\\
\caption{Confusion matrix of MIC on CK+, MMI and AFEW datasets with our MIC.}\label{fig:7} 
\end{figure*}

\noindent\textbf{Results on MMI dataset.} 

The evaluation results on the MMI dataset are shown in Table 3. The performance is also consistent with the CK+ dataset, which evidenced its effectiveness and generality. All of our methods achieve comparable performance to the landmark-based STOA methods. It is more promising that MIC can be significantly better than the methods without the landmarks. Our MIC is efficient, since we explore the correlation in video frames in the compressed domain.

In Figure 6, we give a comparison of accuracy w.r.t. each emotion among five MIC baselines, and the confusion matrix of our MIC is reported in Figure 7 (middle). There is a good performance for the expression class of fear, happiness, sadness, and surprise. In contrast, the accuracy of expression class anger and disgust is relatively limited. Especially, there is a high degree of confusion between anger and disguise. This may be related to the subtle movements between these expressions are relatively in the residual frames.

\noindent\textbf{Results on AFEW dataset.} 

The evaluation of the proposed MIC on the AFEW dataset is shown in Table 4. We note that only the SOTA accuracy obtained by the single-models (non-ensemble model) are listed for a fair comparison. Besides, the audio modality in AFEW can be used to boost the recognition performance \citep{fan2017dynamic,vielzeuf2017temporal,lu2018multiple}. We note that we only focus on the image compression in this paper, and the audio/video data are stored in separate tracks, but the additional modality can also potentially to be added on our framework following the multi-modal methods \citep{fan2017dynamic}.

With the simplified mode variational LSTM-based \citep{baddar2019mode} backbone, the exploration in the compressed domain can achieve comparable or even better recognition performance. More promisingly, our MIC can also achieve real-time processing for the uncontrolled environment, which evidenced its generality. We note that the typical time resolution in FER is 24fps \citep{kim2017multi}.

\begin{table}[t]
	\centering
	\scriptsize
	\begin{tabular}{| l | c | c | c |}
		\hline
		\textbf{Method}   & \textbf{Accuracy} & \textbf{Model type}& \textbf{Ave Test} \\
		\hline \hline

Undirectional LSTM (2017) \citep{vielzeuf2017temporal} & 48.60&Dynamic&-\\
DenseNet-161 (2018) \citep{liu2018multi} & 51.44&Static&-\\

CTSLSTM (2019) \citep{hu2019video} & 51.2 &$\checkmark$&-\\
C3D-GRU (2019) \citep{lee2019visual} & 49.87&Dynamic&-\\
DSTA (2019)$\dag$ \citep{pan2019deep} & 42.98&Dynamic&-\\
E-ConvLSTM (2019)$\dag$ \citep{miyoshi2019facial} & 45.29&Dynamic& 4fps   \\

NST (2020) \citep{kumar2020noisy}$\dag\dag$	 & \textbf{99.69} & Dynamic&-     \\

\hline\hline
Mode VLSTM (2019) \citep{baddar2019mode} & 51.44& Dynamic&  11fps  \\\hline

MIC & \textbf{53.18} & Dynamic&  \textbf{34fps}   \\ 
MIC-MI  & 52.62& Dynamic&   \textbf{34fps}  \\
MIC-MI+Adv\citep{liu2019feature}  & 53.01& Dynamic&  \textbf{34fps}   \\
MIC+$\mathcal{T}^t$  & \textbf{53.18} & Dynamic&   30fps   \\\hline

FAN+ResNet18* (2019) \citep{meng2019frame} &51.18&Static&9fps\\
MIC+ResNet6 & \textbf{53.72} & Dynamic&  \textbf{30fps}   \\ \hline
	\end{tabular}
	\vspace{0.1in}
	\caption{Experimental results on AFEW dataset. *Additional FER+ dataset is used. $\dag$ Optical flow is used. $\dag\dag$Additional body language dataset is used.}
	\label{tab:afew}
\end{table}

In addition, we note that the complementary constraint requires the apex frame in training, which is not applicable for the AFEW dataset. We do not apply the reconstruction loss in the AFEW task. Although the requirement of the apex frame imposes some limitations on the training, it does not affect the generality of the testing or implementation of the trained model with CK+ and MMI.

Some of the works propose to improve the image-based FER networks and combine the frame-wise scores for video-based FER \citep{liu2018multi,meng2019frame}. The image-based FER methods \citep{liu2018multi} achieves high performance, but \citep{liu2018multi} uses a very deep network DenseNet-161 and pretrains it on the private Situ dataset. Moreover, \citep{liu2018multi} utilize the sophisticated post-processing. Actually, an intuition of a statistic-based solution is to avoid LSTM and speed up the processing. However, with the super deep and complicated structure, their processing can be much slower than our solution.  

\citep{vielzeuf2017temporal} uses VGGFace as the backbone of $f_E$ and an RNN model with LSTM units to capture the temporal dynamic cues of the videos. Moreover, \citep{hu2019video,lee2019visual,pan2019deep} also propose to modify the LSTM model for the spatial-temporal modeling. However, all of the above solutions are applied to the decoded space, which requires decoding processing and needs to handle much more complicated data. With the 6-layer ResNet backbone as an expression feature extractor, the performance of our MIC model can be further improved without an additional FER dataset for pre-training. Overall, the proposed MIC can improve the testing speed by a large margin and can achieve the SOTA accuracy as the previous models.

\begin{table}[t]
	\centering
	\footnotesize
	\begin{tabular}{| l | c | c |}
		\hline
		\textbf{Method} & \textbf{Backbone}  & \textbf{Accuracy}  \\
		\hline \hline


Mode variational LSTM \citep{baddar2019mode} & CNN-LSTM & 65.85\%    \\
MIC  & CNN-LSTM &  \textbf{67.28\%}   \\\hline

FAN \citep{meng2019frame} & ResNet18-LSTM & 66.42\%    \\
MIC+FAN  & ResNet18-LSTM &     \textbf{68.53\%}   \\\hline

	\end{tabular}
	\vspace{0.1in}
	\caption{Comparison of the cross-dataset results. We use CK+ training set for training and test on MMI testing set.}
	\label{tab:cross}
\end{table}

\begin{table}[t]
	\centering
	\footnotesize
	\begin{tabular}{| l | c | c |}
		\hline
		\textbf{Method} & \textbf{Backbone}  & \textbf{Accuracy}  \\
		\hline \hline


Mode variational LSTM \citep{baddar2019mode} & CNN-LSTM & 76.32\%    \\
MIC  & CNN-LSTM &  \textbf{80.05\%}   \\\hline

FAN \citep{meng2019frame} & ResNet18-LSTM & 78.75\%    \\
MIC+FAN  & ResNet18-LSTM &     \textbf{81.13\%}   \\\hline

	\end{tabular}
	\vspace{0.1in}
	\caption{Comparison of the cross-dataset results. We use MMI training set for training and test on CK+ testing set.}
	\label{tab:crossb}
\end{table}

\begin{table}[t]
	\centering
	\footnotesize
	\begin{tabular}{| l | c | c |}
		\hline
		\textbf{Method} & \textbf{Backbone}  & \textbf{Accuracy}  \\
		\hline \hline


Mode variational LSTM \citep{baddar2019mode} & CNN-LSTM & 52.34\%    \\
MIC  & CNN-LSTM &  \textbf{55.85\%}   \\\hline

FAN \citep{meng2019frame} & ResNet18-LSTM & 54.72\%    \\
MIC+FAN  & ResNet18-LSTM &     \textbf{56.43\%}   \\\hline

	\end{tabular}
	\vspace{0.1in}
	\caption{Comparison of the cross-dataset results. We use AFEW training set for training and test on MMI testing set for the shared classes.}
	\label{tab:crossc}
\end{table}

\begin{table}[t]
	\centering
	\footnotesize
	\begin{tabular}{| l | c | c |}
		\hline
		\textbf{Method} & \textbf{Backbone}  & \textbf{Accuracy}  \\
		\hline \hline


Mode variational LSTM \citep{baddar2019mode} & CNN-LSTM & 64.60\%    \\
MIC  & CNN-LSTM &  \textbf{67.47\%}   \\\hline

FAN \citep{meng2019frame} & ResNet18-LSTM & 66.29\%    \\
MIC+FAN  & ResNet18-LSTM &     \textbf{68.62\%}   \\\hline

	\end{tabular}
	\vspace{0.1in}
	\caption{Comparison of the cross-dataset results. We use AFEW training set for training and test on CK+ testing set for the shared classes.}
	\label{tab:crossd}
\end{table}

\subsection{Identity feature extraction} 
We adopt the pre-trained face recognizer FaceNet \citep{schroff2015facenet} to extract the identity factor from the I frame. The feature embedded with the convolutional layers and the first fully connected layer can be robust to a broad range of nuisance factors such as expression, pose, illumination, and occlusion variations, since it achieves high accuracy over millions of identities \citep{kemelmacher2016megaface}. To check the expression information in the extracted identity feature $z_I$, we use the feature for expression classification as \cite{liu2019feature}. The results are shown in \ref{tab:id}. We note that achieve zero FER accuracy with $z_I$ does not mean $z_I$ has no information about expression. Instead, approaching the chance probability, i.e., uniform distribution w.r.t. expression classes, indicates the expression is well disentangled from the identity feature $z_I$ with FaceNet.

\begin{table}[t]
	\centering
	\footnotesize
	\begin{tabular}{| l | c | c | c | c | c |}
		\hline
		\textbf{$\beta$} & CK+  & MMI & AFEW    \\
		\hline\hline 
		
$z_I$ & {14.29\%} & 16.68\%  & {14.28\%}  \\  

Chance  & {14.29\%} & 16.67\%  & {14.29\%}  \\\hline
	\end{tabular}
	\vspace{0.1in}
	\caption{Facial expression recognition with the extracted identity feature from I frame.}
	\label{tab:id}
\end{table}

\subsection{Sensitivity analysis}

We use $\alpha$ and $\beta$ to balance the MI and complementary constraint terms and choose the best value with grid searching. In Tab. \ref{tab:sensealpha}, we provide the sensitivity analysis of using different $\alpha$ in three datasets. We can see that we can achieve the best performance on CK+ and MMI with $\alpha=0.1$. For the AFEW dataset, the performance is relatively stable for $\alpha$ from 0.15 to 0.25. We simply use 0.2 for all of our MIC models on the AFEW dataset.

$\beta$ is used to balance the complementary constraint term, which can be helpful for stabilizing the training. In Tab. \ref{tab:sensebeta}, both the fixed $\beta$ and linear changing $\beta$ are compared. Decreasing $\beta$ from 1 to 0 for 30 or 50 epochs can usually achieve the best performance.  

\begin{table*}[t]
	\centering
	\footnotesize
	\begin{tabular}{| l | c | c | c | c | c | c | c | c | c | c | c | c |}
		\hline
		\textbf{$\alpha$} & 0 & 0.01  & 0.05 & 0.1  & 0.15& 0.2  & 0.25& 0.3  & 0.5  & 1 \\
		\hline \hline

CK+  & 97.84\%  & 98.71\% & 98.90\% & \textbf{98.95\%}  & \textbf{98.95\%} & 98.94\% & 98.90\% & 98.88\% & 98.54\% & 98.46\%  \\ \hline

MMI   & 80.25\%  &  81.02\% &  {81.23\%} &\textbf{81.29\%}   & 81.27\% & 81.26\% & 81.13\% & 81.10\%& 81.08\% & 81.04\%    \\\hline

AFEW   & 52.62\%  & 52.92\% & 53.04\% &  53.15\%  & \textbf{53.18\%} & \textbf{53.18\%}& 53.16\% &53.11\% &53.07\% &53.01\%   \\ \hline

	\end{tabular}
	\vspace{0.1in}
	\caption{Sensitivity analysis of hyperparameter $\alpha$ in CK+, MMI and AFEW datasets.}
	\label{tab:sensealpha}
\end{table*}

\begin{table}[t]
	\centering
	\footnotesize
	\begin{tabular}{| l | c | c | c | c | c |}
		\hline
		\textbf{$\beta$} & CK+  & MMI & AFEW    \\
		\hline\hline 
		
Keep 1 & {97.87\%} & 80.69\%  & {52.76\%}  \\  

Keep 0.5 &  {97.90\%} &  {80.98\%} & {52.91\%}  \\ 

Keep 0.1  & {97.92\%} & 81.02\%  & {53.05\%}  \\  

1$\rightarrow$0:30 & \textbf{98.95\%} & \textbf{81.29\%}  & \textbf{53.18\%}   \\

1$\rightarrow$0:50  & \textbf{98.95\%} & 81.26\%  & \textbf{53.18\%}  \\

1$\rightarrow$0:100  & {98.72\%} & 81.17\%  & {53.02\%}  \\

1$\rightarrow$0.5:30  & {97.85\%} & 80.96\%  & {52.88\%}  \\\hline

	\end{tabular}
	\vspace{0.1in}
	\caption{Sensitivity analysis of hyperparameter $\beta$ in CK+, MMI and AFEW datasets. Keep 1 indicates using a fixed $\beta=1$. We denote linear decreasing $\beta$ from 1 to 0 until the 30$^{th}$ epoch as 1$\rightarrow$0:30.}
	\label{tab:sensebeta}
\end{table}

\subsection{Cross-database validation}

Since the subjects are different across CK+ and MMI datasets, the cross-dataset evaluation can be used to evidence if the trained model is affected by identity \cite{meng2017identity}. 
In Tab. \ref{tab:cross}, the FER model trained CK+ training sets in the previous experiment is directly implemented to the MMI testing set. We can see that our MIC can outperform the other methods with the same backbone. In Tab. \ref{tab:crossb}, we use MMI as training data and test on CK+ dataset. The proposed MIC outperforms the other methods with the same backbone consistently.

We note that FAN \cite{meng2019frame} uses ResNet18 as a feature extractor and sequentially pre-trained on MSCeleb-1M face recognition dataset and FER+ expression dataset. 

In Tab. \ref{tab:crossc} and Tab. \ref{tab:crossd}, we use the training set of AFEW for training and test on MMI and CK+, respectively. Considering the large domain shift between AFEW and MMI/CK+, there is a significant performance drop. We note that the proposed method can also achieve better performance than its backbones \cite{baddar2019mode,meng2019frame}.

\subsection{Critical discussion and future work}

The proposed MIC framework has demonstrated its effectiveness w.r.t. accuracy and testing speed. However, MIC highly relies on the separation of I and P frames in compressed video. Although MPEG-4, H.264, and HEVC formats are widely used, some of the compression solutions do not follow the motion compensation with I and P frames. For example, the YUV format compresses the video by considering the different changes of brightness and chromaticity. Moreover, frame loss can be common in real-world video transfer. How will the frame loss affect the FER performance is underexplored.

Identity can be the most challenging variation for FER \cite{liu2017adaptive}, while the FER performance can also be affected by pose and illuminations. It can be promising to take the other variations into account.

The recent work \cite{kumar2020noisy} proposes to utilize the additional unlabeled face dataset to boost the performance, which can be a powerful means for alleviating the scale issue of video-based FER. 

For the large domain gap, e.g., AFEW to MMI/CK+, the domain adaptation methods \cite{he2020image2audio,liu2021subtype} can be used to achieve better cross-dataset performance.

\section{Conclusion}\label{sect:conclusion}

In this paper, we target to explore the facial expression cues directly on the compressed video domain. We are motivated by our practical observation that facial muscle movements can be well encoded in the residual frames, which can be informative and free of cost. Besides, the video compression can reduce the repeating boring patterns in the videos, which rendering the representation to be robust. The increased relevance and reduced complexity or redundancy in FER videos make computation much more effective. We extract the identity and expression factor from the I frame and P frame, respectively, and explicitly enforce their independence with concise and effective mutual information regularization. When the apex frame label is available in training, the complementary constraint can further stabilize the training. In three video-based FER benchmarks, our MIC can improve the performance without the additional identity, face model, or facial landmarks labels. The processing speed of the test stage is promising for real-time FER. Moreover, our mutual information regularization can potentially be a good alternative to adversarial training \citep{liu2019feature} for many disentanglement tasks.

\bibliographystyle{elsarticle-num}

\bibliography{cas-refs}

\begin{thebibliography}{10}
\expandafter\ifx\csname url\endcsname\relax
  \def\url#1{\texttt{#1}}\fi
\expandafter\ifx\csname urlprefix\endcsname\relax\def\urlprefix{URL }\fi
\expandafter\ifx\csname href\endcsname\relax
  \def\href#1#2{#2} \def\path#1{#1}\fi

\bibitem{baddar2019fly}
W.~J. Baddar, S.~Lee, Y.~M. Ro, On-the-fly facial expression prediction using
  lstm encoded appearance-suppressed dynamics, IEEE Transactions on Affective
  Computing (2019).

\bibitem{liu2019permutation}
X.~Liu, Z.~Guo, S.~Li, L.~Kong, P.~Jia, J.~You, B.~Kumar, Permutation-invariant
  feature restructuring for correlation-aware image set-based recognition, in:
  Proceedings of the IEEE International Conference on Computer Vision, 2019,
  pp. 4986--4996.

\bibitem{liu2018adaptive}
X.~Liu, Y.~Ge, C.~Yang, P.~Jia, Adaptive metric learning with deep neural
  networks for video-based facial expression recognition, Journal of Electronic
  Imaging 27~(1) (2018) 013022.

\bibitem{liu2019hard}
X.~Liu, B.~V. Kumar, P.~Jia, J.~You, Hard negative generation for
  identity-disentangled facial expression recognition, Pattern Recognition 88
  (2019) 1--12.

\bibitem{li2020deep}
S.~Li, W.~Deng, Deep facial expression recognition: A survey, IEEE Transactions
  on Affective Computing (2020).

\bibitem{kim2017multi}
D.~H. Kim, W.~J. Baddar, J.~Jang, Y.~M. Ro, Multi-objective based
  spatio-temporal feature representation learning robust to expression
  intensity variations for facial expression recognition, IEEE Transactions on
  Affective Computing 10~(2) (2017) 223--236.

\bibitem{meng2019frame}
D.~Meng, X.~Peng, K.~Wang, Y.~Qiao, frame attention networks for facial
  expression recognition in videos, in: 2019 IEEE International Conference on
  Image Processing (ICIP), IEEE, 2019, pp. 3866--3870.

\bibitem{liu2017adaptive}
X.~Liu, B.~{Vijaya Kumar}, J.~You, P.~Jia, Adaptive deep metric learning for
  identity-aware facial expression recognition, in: CVPRW, 2017, pp. 522--531.

\bibitem{kahou2016emonets}
S.~E. Kahou, X.~Bouthillier, P.~Lamblin, C.~Gulcehre, V.~Michalski, K.~Konda,
  S.~Jean, P.~Froumenty, Y.~Dauphin, N.~Boulanger-Lewandowski, et~al., Emonets:
  Multimodal deep learning approaches for emotion recognition in video, Journal
  on Multimodal User Interfaces 10~(2) (2016) 99--111.

\bibitem{xu2016video}
B.~Xu, Y.~Fu, Y.-G. Jiang, B.~Li, L.~Sigal, Video emotion recognition with
  transferred deep feature encodings, in: Proceedings of the 2016 ACM on
  International Conference on Multimedia Retrieval, 2016, pp. 15--22.

\bibitem{yeo2006compressed}
C.~Yeo, P.~Ahammad, K.~Ramchandran, S.~S. Sastry, Compressed domain real-time
  action recognition, in: 2006 IEEE Workshop on Multimedia Signal Processing,
  IEEE, 2006, pp. 33--36.

\bibitem{richardson2003h264}
I.~E. Richardson, H264/mpeg-4 part 10 white paper-prediction of intra
  macroblocks, Internet Citation, Apr 30 (2003).

\bibitem{le1991mpeg}
D.~Le~Gall, Mpeg: A video compression standard for multimedia applications,
  Communications of the ACM 34~(4) (1991) 46--58.

\bibitem{lucey2010extended}
P.~Lucey, J.~F. Cohn, T.~Kanade, J.~Saragih, Z.~Ambadar, I.~Matthews, The
  extended cohn-kanade dataset (ck+): A complete dataset for action unit and
  emotion-specified expression, in: 2010 ieee computer society conference on
  computer vision and pattern recognition-workshops, IEEE, 2010, pp. 94--101.

\bibitem{wu2018compressed}
C.-Y. Wu, M.~Zaheer, H.~Hu, R.~Manmatha, A.~J. Smola, P.~Kr{\"a}henb{\"u}hl,
  Compressed video action recognition, in: Proceedings of the IEEE Conference
  on Computer Vision and Pattern Recognition, 2018, pp. 6026--6035.

\bibitem{meng2017identity}
Z.~Meng, P.~Liu, J.~Cai, S.~Han, Y.~Tong, Identity-aware convolutional neural
  network for facial expression recognition, in: 2017 12th IEEE International
  Conference on Automatic Face \& Gesture Recognition (FG 2017), IEEE, 2017,
  pp. 558--565.

\bibitem{calder2005understanding}
A.~J. Calder, A.~W. Young, Understanding the recognition of facial identity and
  facial expression, Nature Reviews Neuroscience 6~(8) (2005) 641--651.

\bibitem{ali2019all}
K.~Ali, C.~E. Hughes, All-in-one: Facial expression transfer, editing and
  recognition using a single network, arXiv preprint arXiv:1911.07050 (2019).

\bibitem{cai2019identity}
J.~Cai, Z.~Meng, A.~S. Khan, Z.~Li, J.~O'Reilly, Y.~Tong, Identity-free facial
  expression recognition using conditional generative adversarial network,
  arXiv preprint arXiv:1903.08051 (2019).

\bibitem{dhall2014emotion}
A.~Dhall, R.~Goecke, J.~Joshi, K.~Sikka, T.~Gedeon, Emotion recognition in the
  wild challenge 2014: Baseline, data and protocol, in: Proceedings of the 16th
  international conference on multimodal interaction, 2014, pp. 461--466.

\bibitem{schroff2015facenet}
F.~Schroff, D.~Kalenichenko, J.~Philbin, Facenet: A unified embedding for face
  recognition and clustering, in: Proceedings of the IEEE conference on
  computer vision and pattern recognition, 2015, pp. 815--823.

\bibitem{kemelmacher2016megaface}
I.~Kemelmacher-Shlizerman, S.~M. Seitz, D.~Miller, E.~Brossard, The megaface
  benchmark: 1 million faces for recognition at scale, in: Proceedings of the
  IEEE conference on computer vision and pattern recognition, 2016, pp.
  4873--4882.

\bibitem{liu2019feature}
X.~Liu, S.~Li, L.~Kong, W.~Xie, P.~Jia, J.~You, B.~{Vijaya Kumar},
  Feature-level frankenstein: Eliminating variations for discriminative
  recognition, in: Proceedings of the IEEE Conference on Computer Vision and
  Pattern Recognition, 2019, pp. 637--646.

\bibitem{liu2021mutual}
X.~Liu, Y.~Chao, J.~J. You, C.-C.~J. Kuo, B.~Vijayakumar, Mutual information
  regularized feature-level frankenstein for discriminative recognition, IEEE
  Transactions on Pattern Analysis and Machine Intelligence (2021).

\bibitem{linsker1988self}
R.~Linsker, Self-organization in a perceptual network, Computer 21~(3) (1988)
  105--117.

\bibitem{belghazi2018mine}
M.~I. Belghazi, A.~Baratin, S.~Rajeswar, S.~Ozair, Y.~Bengio, A.~Courville,
  R.~D. Hjelm, Mine: mutual information neural estimation, arXiv preprint
  arXiv:1801.04062 (2018).

\bibitem{goodfellow2016nips}
I.~Goodfellow, Nips 2016 tutorial: Generative adversarial networks, arXiv
  preprint arXiv:1701.00160 (2016).

\bibitem{liu2021identity}
X.~Liu, L.~Jin, X.~Han, J.~Lu, J.~You, L.~Kong, Identity-aware facial
  expression recognition in compressed video, ICPR (2020).

\bibitem{Jiyoung2020multi}
L.~Jiyoung, K.~Sunok, K.~Seungryong, S.~Kwanghoon, Multi-modal recurrent
  attention networks for facial expression recognition, IEEE Transactions on
  Image Processing (2020).

\bibitem{sandbach2012static}
G.~Sandbach, S.~Zafeiriou, M.~Pantic, L.~Yin, Static and dynamic 3d facial
  expression recognition: A comprehensive survey, Image and Vision Computing
  30~(10) (2012) 683--697.

\bibitem{al2018deep}
D.~A. AL~CHANTI, A.~Caplier, Deep learning for spatio-temporal modeling of
  dynamic spontaneous emotions, IEEE Transactions on Affective Computing
  (2018).

\bibitem{donahue2015long}
J.~Donahue, L.~Anne~Hendricks, S.~Guadarrama, M.~Rohrbach, S.~Venugopalan,
  K.~Saenko, T.~Darrell, Long-term recurrent convolutional networks for visual
  recognition and description, in: Proceedings of the IEEE conference on
  computer vision and pattern recognition, 2015, pp. 2625--2634.

\bibitem{jain2017multi}
D.~K. Jain, Z.~Zhang, K.~Huang, Multi angle optimal pattern-based deep learning
  for automatic facial expression recognition, Pattern Recognition Letters
  (2017).

\bibitem{barros2016developing}
P.~Barros, S.~Wermter, Developing crossmodal expression recognition based on a
  deep neural model, Adaptive behavior 24~(5) (2016) 373--396.

\bibitem{zhao2018learning}
J.~Zhao, X.~Mao, J.~Zhang, Learning deep facial expression features from image
  and optical flow sequences using 3d cnn, The Visual Computer 34~(10) (2018)
  1461--1475.

\bibitem{zhang2018real}
B.~Zhang, L.~Wang, Z.~Wang, Y.~Qiao, H.~Wang, Real-time action recognition with
  deeply transferred motion vector cnns, IEEE Transactions on Image Processing
  27~(5) (2018) 2326--2339.

\bibitem{zhang2016real}
B.~Zhang, L.~Wang, Z.~Wang, Y.~Qiao, H.~Wang, Real-time action recognition with
  enhanced motion vector cnns, in: Proceedings of the IEEE conference on
  computer vision and pattern recognition, 2016, pp. 2718--2726.

\bibitem{hyvarinen1999nonlinear}
A.~Hyv{\"a}rinen, P.~Pajunen, Nonlinear independent component analysis:
  Existence and uniqueness results, Neural Networks 12~(3) (1999) 429--439.

\bibitem{brakel2017learning}
P.~Brakel, Y.~Bengio, Learning independent features with adversarial nets for
  non-linear ica, arXiv preprint arXiv:1710.05050 (2017).

\bibitem{hjelm2018learning}
R.~D. Hjelm, A.~Fedorov, S.~Lavoie-Marchildon, K.~Grewal, P.~Bachman,
  A.~Trischler, Y.~Bengio, Learning deep representations by mutual information
  estimation and maximization, arXiv preprint arXiv:1808.06670 (2018).

\bibitem{velivckovic2018deep}
P.~Veli{\v{c}}kovi{\'c}, W.~Fedus, W.~L. Hamilton, P.~Li{\`o}, Y.~Bengio, R.~D.
  Hjelm, Deep graph infomax, arXiv preprint arXiv:1809.10341 (2018).

\bibitem{liu2018exploring}
Y.~Liu, F.~Wei, J.~Shao, L.~Sheng, J.~Yan, X.~Wang, Exploring disentangled
  feature representation beyond face identification, in: Proceedings of the
  IEEE Conference on Computer Vision and Pattern Recognition, 2018, pp.
  2080--2089.

\bibitem{hadad2018two}
N.~Hadad, L.~Wolf, M.~Shahar, A two-step disentanglement method, in:
  Proceedings of the IEEE Conference on Computer Vision and Pattern
  Recognition, 2018, pp. 772--780.

\bibitem{baddar2019mode}
W.~J. Baddar, Y.~M. Ro, Mode variational lstm robust to unseen modes of
  variation: Application to facial expression recognition, in: Proceedings of
  the AAAI Conference on Artificial Intelligence, Vol.~33, 2019, pp.
  3215--3223.

\bibitem{kumawat2019lbvcnn}
S.~Kumawat, M.~Verma, S.~Raman, Lbvcnn: Local binary volume convolutional
  neural network for facial expression recognition from image sequences, in:
  Proceedings of the IEEE Conference on Computer Vision and Pattern Recognition
  Workshops, 2019, pp. 0--0.

\bibitem{xie2017adversarial}
Q.~Xie, Z.~Dai, Y.~Du, E.~Hovy, G.~Neubig, Controllable invariance through
  adversarial feature learning, in: NIPS, 2017, pp. 585--596.

\bibitem{kanade2000comprehensive}
T.~Kanade, J.~F. Cohn, Y.~Tian, Comprehensive database for facial expression
  analysis, in: Proceedings Fourth IEEE International Conference on Automatic
  Face and Gesture Recognition (Cat. No. PR00580), IEEE, 2000, pp. 46--53.

\bibitem{pantic2005web}
M.~Pantic, M.~Valstar, R.~Rademaker, L.~Maat, Web-based database for facial
  expression analysis, in: 2005 IEEE international conference on multimedia and
  Expo, IEEE, 2005, pp. 5--pp.

\bibitem{paninski2003estimation}
L.~Paninski, Estimation of entropy and mutual information, Neural computation
  15~(6) (2003) 1191--1253.

\bibitem{zhang2017facial}
K.~Zhang, Y.~Huang, Y.~Du, L.~Wang, Facial expression recognition based on deep
  evolutional spatial-temporal networks, IEEE Transactions on Image Processing
  26~(9) (2017) 4193--4203.

\bibitem{lee2019visual}
M.~K. Lee, D.~Y. Choi, D.~H. Kim, B.~C. Song, Visual scene-aware hybrid neural
  network architecture for video-based facial expression recognition, in: 2019
  14th IEEE International Conference on Automatic Face \& Gesture Recognition
  (FG 2019), IEEE, 2019, pp. 1--8.

\bibitem{hu2019video}
M.~Hu, H.~Wang, X.~Wang, J.~Yang, R.~Wang, Video facial emotion recognition
  based on local enhanced motion history image and cnn-ctslstm networks,
  Journal of Visual Communication and Image Representation 59 (2019) 176--185.

\bibitem{verma2019facial}
M.~Verma, H.~Kobori, Y.~Nakashima, N.~Takemura, H.~Nagahara, Facial expression
  recognition with skip-connection to leverage low-level features, in: 2019
  IEEE International Conference on Image Processing (ICIP), IEEE, 2019, pp.
  51--55.

\bibitem{albrici2019g2}
T.~Albrici, M.~Fasounaki, S.~B. Salimi, G.~Vray, B.~Bozorgtabar, H.~K. Ekenel,
  J.-P. Thiran, G2-ver: Geometry guided model ensemble for video-based facial
  expression recognition, in: 2019 14th IEEE International Conference on
  Automatic Face \& Gesture Recognition (FG 2019), IEEE, 2019, pp. 1--6.

\bibitem{kumar2020noisy}
V.~Kumar, S.~Rao, L.~Yu, Noisy student training using body language dataset
  improves facial expression recognition, in: European Conference on Computer
  Vision, Springer, 2020, pp. 756--773.

\bibitem{perveen2018spontaneous}
N.~Perveen, D.~Roy, C.~K. Mohan, Spontaneous expression recognition using
  universal attribute model, IEEE Transactions on Image Processing 27~(11)
  (2018) 5575--5584.

\bibitem{dhall2017individual}
A.~Dhall, R.~Goecke, S.~Ghosh, J.~Joshi, J.~Hoey, T.~Gedeon, From individual to
  group-level emotion recognition: Emotiw 5.0, in: ICMI, 2017, pp. 524--528.

\bibitem{he2016deep}
K.~He, X.~Zhang, S.~Ren, J.~Sun, Deep residual learning for image recognition,
  in: CVPR, 2016, pp. 770--778.

\bibitem{liu2014deeply}
M.~Liu, S.~Li, S.~Shan, R.~Wang, X.~Chen, Deeply learning deformable facial
  action parts model for dynamic expression analysis, in: Asian conference on
  computer vision, Springer, 2014, pp. 143--157.

\bibitem{shou2019dmc}
Z.~Shou, X.~Lin, Y.~Kalantidis, L.~Sevilla-Lara, M.~Rohrbach, S.-F. Chang,
  Z.~Yan, Dmc-net: Generating discriminative motion cues for fast compressed
  video action recognition, in: CVPR, 2019, pp. 1268--1277.

\bibitem{fan2017dynamic}
X.~Fan, T.~Tjahjadi, A dynamic framework based on local zernike moment and
  motion history image for facial expression recognition, Pattern recognition
  64 (2017) 399--406.

\bibitem{vielzeuf2017temporal}
V.~Vielzeuf, S.~Pateux, F.~Jurie, Temporal multimodal fusion for video emotion
  classification in the wild, in: ICMI, 2017, pp. 569--576.

\bibitem{lu2018multiple}
C.~Lu, W.~Zheng, C.~Li, C.~Tang, S.~Liu, S.~Yan, Y.~Zong, Multiple
  spatio-temporal feature learning for video-based emotion recognition in the
  wild, in: Proceedings of the 20th ACM International Conference on Multimodal
  Interaction, 2018, pp. 646--652.

\bibitem{liu2018multi}
C.~Liu, T.~Tang, K.~Lv, M.~Wang, Multi-feature based emotion recognition for
  video clips, in: ICMI, 2018, pp. 630--634.

\bibitem{pan2019deep}
X.~Pan, G.~Ying, G.~Chen, H.~Li, W.~Li, A deep spatial and temporal aggregation
  framework for video-based facial expression recognition, IEEE Access 7 (2019)
  48807--48815.

\bibitem{miyoshi2019facial}
R.~Miyoshi, N.~Nagata, M.~Hashimoto, Facial-expression recognition from video
  using enhanced convolutional lstm, in: 2019 Digital Image Computing:
  Techniques and Applications (DICTA), IEEE, 2019, pp. 1--6.

\bibitem{he2020image2audio}
G.~He, X.~Liu, F.~Fan, J.~You, Image2audio: Facilitating semi-supervised audio
  emotion recognition with facial expression image, in: CVPR Workshops, 2020,
  pp. 912--913.

\bibitem{liu2021subtype}
X.~Liu, X.~Liu, B.~Hu, W.~Ji, F.~Xing, J.~Lu, J.~You, C.-C.~J. Kuo, G.~E.
  Fakhri, J.~Woo, Subtype-aware unsupervised domain adaptation for medical
  diagnosis, AAAI (2021).

\end{thebibliography}

\end{document}